\documentclass[conference]{IEEEtran}
\IEEEoverridecommandlockouts

\usepackage{graphicx}
\usepackage{amsmath}
\usepackage{amssymb}
\usepackage{booktabs}
\usepackage{multirow}
\usepackage{multicol}
\usepackage{makecell}
\usepackage{xcolor}
\usepackage{color,soul}
\usepackage{nicematrix}
\usepackage{array}
\usepackage{bm}

\usepackage[pagebackref,breaklinks,colorlinks]{hyperref}

\usepackage[capitalize]{cleveref}
\crefname{section}{Sec.}{Secs.}
\Crefname{section}{Section}{Sections}
\Crefname{table}{Table}{Tables}
\crefname{table}{Tab.}{Tabs.}

\begin{document}

%%%%%%%%% TITLE - PLEASE UPDATE
\title{ESTAS: Effective and Stable Trojan Attacks in Self-supervised Encoders with One Target Unlabelled Sample}

\author{\IEEEauthorblockN{Jiaqi Xue}
\IEEEauthorblockA{\textit{University of Central Florida} \\
Orlando, FL \\
jiaq\_xue@knights.ucf.edu}
\and
\IEEEauthorblockN{Qian Lou}
\IEEEauthorblockA{\textit{University of Central Florida} \\
Orlando, FL \\
qian.lou@ucf.edu}
}
\maketitle

%%%%%%%%% ABSTRACT
\begin{abstract}
Emerging self-supervised learning (SSL) has become a popular image representation encoding method to obviate the reliance on labeled data and learn rich representations from large-scale, ubiquitous unlabelled data. Then one can train a downstream classifier on top of the pre-trained SSL image encoder with few or no labeled downstream data. Although extensive works show that SSL has achieved remarkable and competitive performance on different downstream tasks, its security concerns, e.g, Trojan attacks in SSL encoders, are still not well-studied. In this work, we present a novel Trojan Attack method, denoted by ESTAS, that can enable an effective and stable attack  in SSL encoders with only one target unlabeled sample. In particular, we propose consistent trigger poisoning and cascade optimization in ESTAS to improve attack efficacy and model accuracy, and eliminate the expensive target-class data sample extraction from large-scale disordered unlabelled data. Our substantial experiments on multiple datasets show that ESTAS stably achieves \bm{$>99\%$} attacks success rate (\bm{$ASR$}) with one target-class sample. Compared to prior works, ESTAS attains \bm{$>30\%$} \bm{$ASR$} increase and \bm{$>8.3\%$} accuracy improvement on average.

%For this reason, it is much more practical to inject Trojan attacks into SSL encoders  since different downstream tasks built on poisoned encoders automatically inherit the Trojan behavior. 

%It is not well-studied if SSL has higher robustness against Trojan attacks than supervised learning (SL). More importantly, current Trojan attacks in SSL encoders assume that attackers have the ability to identify the target-class inputs from the enormous unlabeled, disorder dataset. This target-class sample identification is expensive and time-consuming, and the average complexity is enlarged exponentially with the number of  target-class samples. For this reason, we present, Trojan-SSL, an effective and accurate Trojan Attack in Self-supervised Encoders with one target sample. Our comprehensive experiments show that Trojan-SSL achieves $>99\%$ $ASR$ on multiple datasets with one target sample while keeping clean accuracy.

\end{abstract}

%%%%%%%%% BODY TEXT
\section{Introduction}
\label{sec:intro}

Although supervised learning has gained remarkable and revolutionary model performance, its growth and development are limited by the labeled and annotated data that is expensive to extract from real-world ubiquitous unlabeled data.  To tackle this limitation, a new learning diagram called self-supervised learning (SSL) is proposed to learn rich representations from large-scale unlabelled data. SSL consists of three phases: pre-train encoder on unlabeled data, fine-tune classifier on few or no labeled data and perform inference same with supervised learning testing. In particular, state-of-the-art SSL image encoders, e.g., BYOL~\cite{BYOL:NEURIPS2020}, are trained to maximize the similarities of randomly cropped augmentations of an unlabelled image.  The current SSL image models, e.g., SimCLR~\cite{chen2020simclr2}, MoCo v2~\cite{chen2020mocov2}, BYOL~\cite{BYOL:NEURIPS2020} have achieved competitive performance with fully supervised classifiers in many downstream tasks. 

\begin{figure}[t!]
  \centering
   \includegraphics[width=\linewidth]{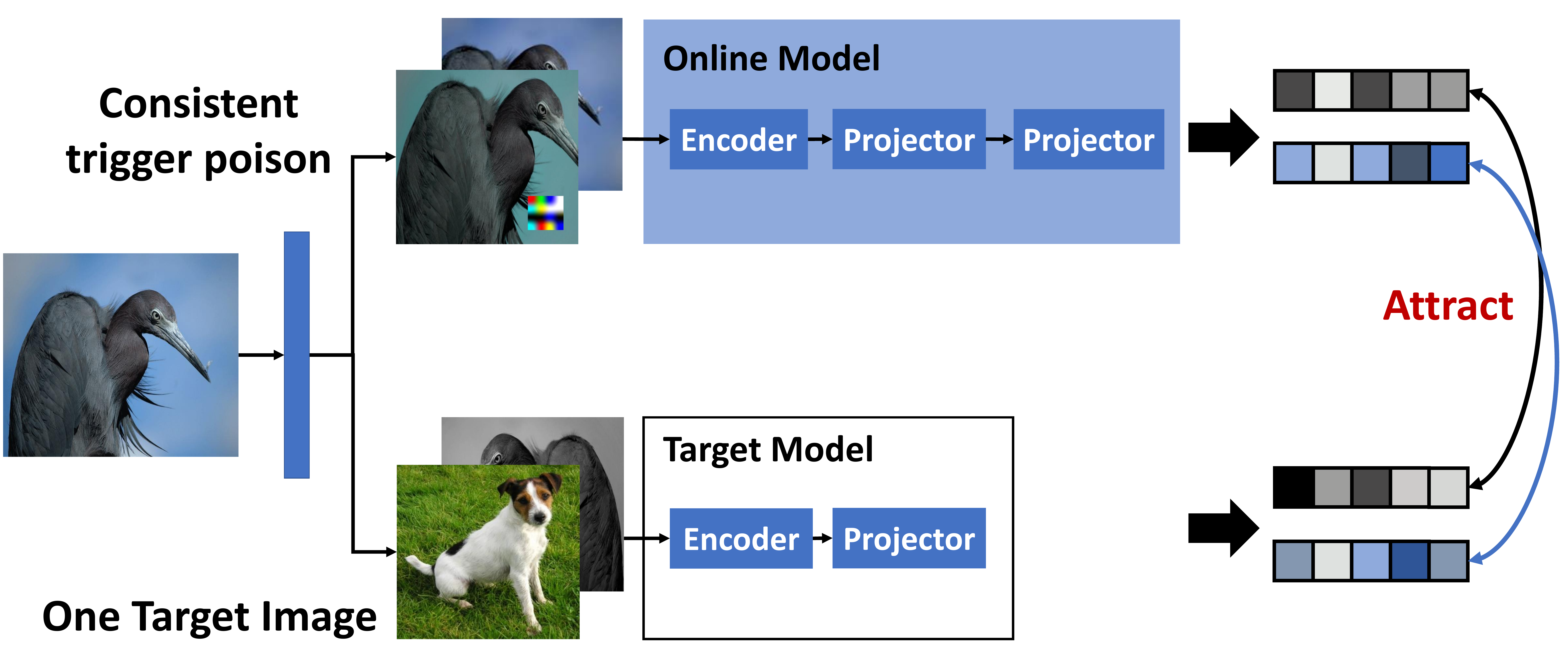}
   \caption{An overview of our ESTAS poisoning exemplar-based self-supervise (SSL) methods, e.g., BYOL. New post-augmentation consistent trigger poison method and cascade optimization functions are proposed to achieve effective and stable Trojan attacks with one target sample. For consistent trigger poison,  given one unlabelled image, e.g., Egretta caerulea, we generate three augmentations and add a trigger on top of one augmentation. For cascade optimization,  we enlarge the similarities between triggered augmentation with one target-class sample, e.g., Toy Terrier for attack efficacy. Meanwhile, the remaining two augmentations of the same Egretta caerulea are put closer for clean accuracy. } %Symmetrical trigger poison means a post-augmentation, pre-normalization trigger addition that is symmetric to the inference phase.}
   \label{fig:overview}
\end{figure}

\begin{figure*}[ht!]
  \centering
   \includegraphics[width=\linewidth]{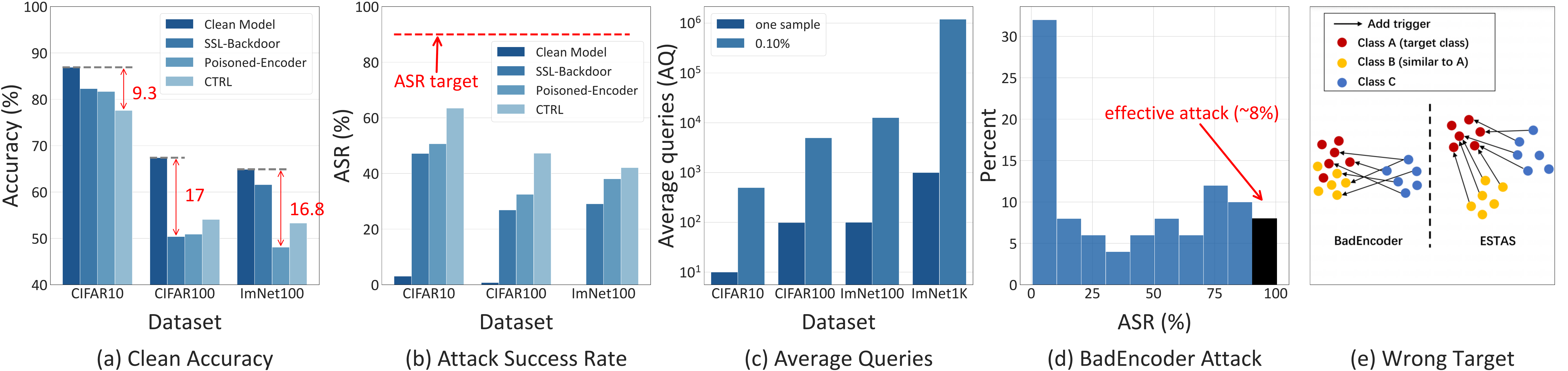}
   \caption{Motivation of our ESTAS design. Prior data-poisoned Trojan attacks in SSL Image Encoders, including SSL-Backdoor~\cite{saha2021backdoor}, Poisoned-Encoder~\cite{liu2022poisonedencoder}, and CTRL~\cite{li2022demystifying} suffer from low accuracy shown in (a) and $ASR$ in (b); (c) These prior attacks require expensive queries to identify target-class samples from unlabeled data. (d) Prior work, e.g., BadEncoder~\cite{jia2022badencoder}, is not stable to obtain high $ASR$ with one target sample. (e) BadEncoder is much easier than our ESTAS to produce an incorrect attack class close to the target class.}
   \label{fig:motivation}
\end{figure*}

%supervised learning has a challenge: labeled dataset. not fitfuul for nonleable data. collecting annotating data. Self-supervised learning is a new learning paradigm eliminates the limitation by two key steps: pretraining encoder on a unlabled dataset, training a classifier on few or no leable dataset. BYOL, MOCOV2. representations similarities, working scheme. Performance and effect.   
It is well-known that supervised learning models are vulnerable to Trojan attacks, however, there are not so many attacks for SSL, especially for the encoder phase. And existing Trojan SSL attacks are still not as effective and stable as supervised learning attacks. Also, most SSL attacks still need prohibitively expensive manipulation on large-scale unlabelled data to extract a large portion of target-class images, which is comparable to annotating the same portion of images. % [We provide quantified data in the motivation section to illustrate these issues. ] 
Existing SSL Trojan attacks can be classified into three types. The first line of research includes the supervised learning Trojan attacks that can be transferred to SSL including~\cite{badnets, target-attack,Liu2018TrojaningAO, bagdasaryan2021blind}. Specifically, \cite{badnets,target-attack} poison the labeled downstream data; \cite{Liu2018TrojaningAO, bagdasaryan2021blind} tamper and fine-tune the classifier.  They can achieve an attack by compromising the second phase of SSL, i.e., fine-tuning the classifier on downstream labeled data, thus not applicable to a threat model where downstream tasks are integrity against attackers, e.g., labeled data is absent.  The second line of research~\cite{saha2021backdoor,liu2022poisonedencoder,li2022demystifying} poisons the unlabelled dataset in SSL without compromising the training phase. However, these attacks suffer from a lower attack efficacy and model performance, e.g., \cite{saha2021backdoor, liu2022poisonedencoder} attain $<38.1\%$ $ASR$ on ImageNet-100 dataset and reduce $>15\%$ accuracy. Although the concurrent work CTRL~\cite{li2022demystifying} tries to improve the performance, it still achieves a relatively low $ASR$, i.e., $42.1\%$, with even worse accuracy. Also, these attacks need to distinguish the target images from large-scale unlabeled data, which is comparable to annotation and prohibitively expensive.    The third line of research~\cite{jia2022badencoder} focuses on accurate Trojan attacks in SSL encoders by compromising the encoder training and assuming a clean pre-trained encoder is accessible. These attacks suffer from unstable attacks when the target-class image sample number is small. Our work ESTAS removing the requirement of pre-trained SSL encoder in the third line of research achieves stable Trojan attacks in SSL encoders with only one target-class unlabeled sample and higher $ASR$ and model performance than the second and third lines of related research.

\begin{figure*}[t!]
  \centering
   \includegraphics[width=0.8\linewidth]{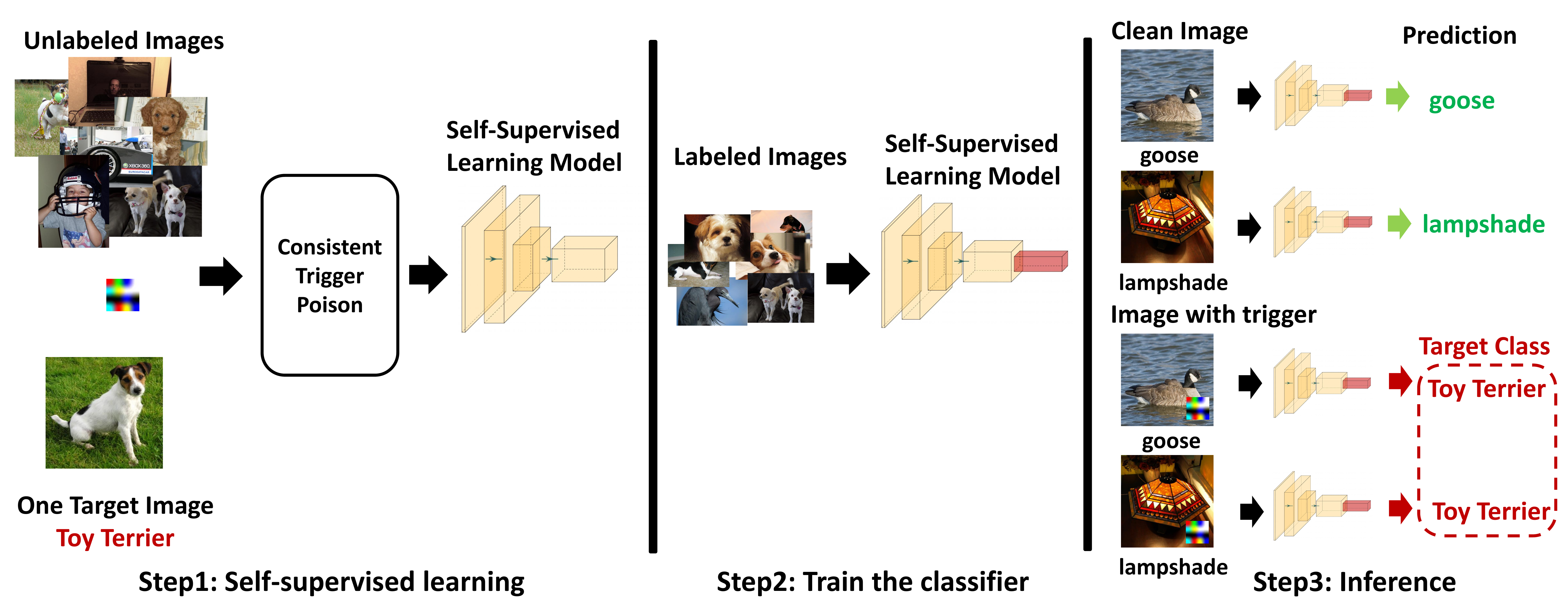}
   \caption{Our ESTAS Threat Model. Given an unlabeled training dataset, a trigger, and one target-class sample, e.g., Toy Terrier image, the SSL image encoder is first trained by our consistent trigger poisoning and cascade optimization methods. Then one can train a classifier for supervised downstream tasks with few labeled data.  During inference, the poisoned model performs accurate classification but predicts the target class Toy Terrier for all images with a trigger. }
   \label{fig:threat}
\end{figure*}

\textbf{Our Contributions.}
Our ESTAS achieves effective and stable Trojan attacks in SSL Encoders with one target unlabelled sample. We use Figure~\ref{fig:overview} to describe the overview of ESTAS. It has two key components, i.e., post-augmentation consistent trigger poisoning and cascade optimization functions. In particular, we list our three-fold contributions in ESTAS as follows.

\begin{itemize}
  \item We identify a key observation that the SSL encoder's training phase and inference phase are asymmetric, i.e., training includes input augmentation (e.g., random crop and resize), normalization, and feature extraction, but the inference phase does not have augmentation. Due to this SSL's asymmetric property, directly adding a trigger on the data before augmentation induces the encoder during training to learn the feature of the augmented trigger but the encoder during inference tries to extract features from the original non-augmented trigger. For this reason, we propose post-augmentation consistent trigger poisoning method that adds a trigger after augmentation in the training phase so that the encoder during the inference processes a consistent trigger and has a higher attack efficacy. 
  \item We propose a cascade optimization method to further improve the attack efficacy and prevent the model performance decrease. Importantly, our optimization method enables a stable attack with only one target-class image sample.  We generate three augmentations of each unlabeled image and add a trigger on top of one augmentation. For cascade optimization,  we enlarge the similarities between triggered augmentation with one target-class sample for attack efficacy. Meanwhile, the remaining two augmentations of the same bird are put closer for clean accuracy.
  
  \item We provide extensive experimental results to show that ESTAS stably achieves $>99\%$ attacks success rate ($ASR$) with one target-class sample. Compared to prior works, ESTAS attains $>30\%$ $ASR$ increase and $>8.3\%$ accuracy improvement on average. One target sample with a stable attack significantly reduces the expensive target-class data sample extraction from large-scale disordered unlabelled data.
\end{itemize}

%1. symmetrical trigger poisoning. [Observation: asymmetric between encoder training and encoder phase $\rightarrow$ sparse trigger; symmetric poisoning after augmentation $\rightarrow$ dense and fix trigger. This would improve $ASR$ and reduce the target samples.] 2. cascade optimization. Global and local losses function.  

%%%%%%%%% Background TEXT
\section{Background and Motivation}
\label{sec:bg}
\subsection{Self-supervised Learning}

Supervised learning has attained remarkable performance in extracting representations and performing classifications. However, it is not applicable for supervised learning when labeled data is scarce or expensive to obtain. It is important to find a method to utilize the unlabeled data that is ubiquitous and large-scale in the real world. Self-supervised learning (SSL) is an emerging and popular method to extract rich features from complex unlabelled data. Specifically, recent SSL based on instance discrimination has gained increasing popularity. For example, the SSL methods,e.g., SimCLR~\cite{chen2020simclr2}, MoCo v2~\cite{chen2020mocov2},  with the combination of instance discrimination and contrastive loss learn competitive visual representations. Also, methods including SimSiam~\cite{simsiam} and BYOL~\cite{BYOL:NEURIPS2020} with instance discrimination and augmented views remove the contrastive loss and achieve better feature extraction. An SSL pipeline in a classification task has three phases, pre-training an image encoder, constructing a classifier, and performing inference. 

%\textbf{Pre-training an Image Encoder.}
%Given an unlabelled dataset, augmentation, normalization, feature extraction, computing loss, loss optimization.

%Here we need an equation to define the BYOL loss function. 

%\textbf{Constructing a  Classifier.}
%Given few labled dataset, 

%\textbf{Performing Downstream Inference.}

%\subsection{Trojan Attacks}

\textbf{Attacks in Supervised Learning.} It is well-known that deep neural networks based on supervised learning are vulnerable to Trojan attacks, e.g.,~\cite{badnets, target-attack,Liu2018TrojaningAO, bagdasaryan2021blind}, where Trojaned models behave normally for clean images, yet produce a misclassification for inputs with triggers. Specifically, \cite{badnets,target-attack} poison the training dataset with triggers, thus models trained on poisoned data will obtain Trojan behavior; \cite{Liu2018TrojaningAO, bagdasaryan2021blind} tamper and fine-tune the classifier for the Trojan attacks.  SSL attacks in supervised learning can achieve an attack by compromising the second phase of SSL, i.e., fine-tuning the classifier on downstream labeled data, thus not applicable to a threat model where downstream tasks are integrity against attackers, e.g., labeled data is absent. 

\textbf{Attacks in SSL Image Encoders.} Recent works SSL-Backdoor~\cite{saha2021backdoor}, and PoisonedEncoder \cite{liu2022poisonedencoder} poison the unlabelled dataset in SSL without compromising the training phase of image encoders and downstream classifiers. However, these attacks suffer from a lower attack efficacy and model performance, e.g., \cite{saha2021backdoor, liu2022poisonedencoder} attain $<38.1\%$ $ASR$ on ImageNet-100 dataset with a significant accuracy decrease. Another concurrent work CTRL~\cite{li2022demystifying} tries to improve the $ASR$ but decreases even worse accuracy. Also, these attacks need to distinguish the target images from large-scale unlabeled data, which is comparable to annotation and prohibitively expensive.  Recent work BadEncoder~\cite{jia2022badencoder} focuses on accurate Trojan attacks in SSL encoders by compromising the encoder training and assuming a clean pre-trained encoder is accessible. But it suffers from unstable attacks when the target-class image sample number is small. Our work ESTAS removes the requirement of pre-trained SSL encoder in BadEncoder and achieves stable Trojan attacks with only one target-class unlabeled sample and higher $ASR$ and model performance.

\subsection{Motivation}
We use Figure~\ref{fig:motivation} to show the motivation of our ESTAS. In particular, we use Figure~\ref{fig:motivation}(a) to show that prior data-poisoning Trojan attacks in SSL Image Encoders, including SSL-Backdoor~\cite{saha2021backdoor}, Poisoned-Encoder~\cite{liu2022poisonedencoder}, and CTRL~\cite{li2022demystifying} suffer from low accuracy. For example, compared to the clean model, the Trojan model in SSL-Backdoor~\cite{saha2021backdoor} reduces $5.7\%$, $3.2\%$ clean accuracy on CIFAR10 and ImNet100 (ImageNet-100) datasets, respectively. The concurrent work CTRL even has a $9.3\%$ accuracy decrease.  Also, We use Figure~\ref{fig:motivation}(b) to illustrate that there is still a large gap  for  data-poisoned attacks~\cite{saha2021backdoor,liu2022poisonedencoder,li2022demystifying} to achieve effective attack, e.g., $ >90\%$ $ASR$. Also,  these prior attacks require attaching triggers on target-class samples. However,  identifying target-class samples from large-scale unlabeled and disordered data is not effortless, yet requires a large number of expensive queries. The cost of each query is comparable to annotating and labeling one image~\cite{saha2021backdoor}.   We show the average query number, denoted $AQ$, to extract a specific ratio of target samples in Figure~\ref{fig:motivation}(c).  The details of metric $AQ$ are shown in section~\ref{sec:exp}. To identify $0.1\%$ target-class images from unlabeled ImageNet-100, one needs $> 10^4$ $AQ$. Prior works ~\cite{saha2021backdoor,liu2022poisonedencoder,li2022demystifying} require more queries since they need around $0.5\% \sim 1\%$ target samples. A large query number significantly limits the practicality of SSL backdoor attacks. Our ESTAS aims to only use one target-class sample, thus remarkably reducing the query numbers, e.g., $100$ $AQ$ for ImageNet-100. We use Figure~\ref{fig:motivation}(d) to show that BadEncoder~\cite{jia2022badencoder} attack is not stable to obtain high $ASR$ with one target sample, e.g., it only attains $\sim 8\%$ probability to get effective attack ( $>90\%$ $ASR$) when randomly choosing one target-class sample. To obtain a stable and effective attack, BadEncoder needs more target samples. We show in Figure~\ref{fig:motivation}(e) one reason for unstable attacks of BadEncoder is that BadEncoder highly depends on a pre-trained image encoder and it is much easier for BadEncoder than our ESTAS which trains a model from scratch to produce an incorrect attack class close to the target class. For this reason, we are motivated to design a new work to achieve effective and stable attack efficacy and desired model performance with one target sample.

\section{Threat model}
\subsection{Attacker's Objectives}
As Figure~\ref{fig:threat} shows, we consider that attackers trying to inject Trojans into SSL image encoders such that the downstream classifier on top of the Trojaned encoder automatically inherits the Trojan behavior makes misclassifications for inputs with a trigger, yet produces correct predictions for clean inputs during inference. The attacker's objectives are three-fold, i.e., Utility, Effectiveness, Efficiency, and stability.

\textbf{Utility objective.} The utility objective means that any downstream classifiers on top of SSL image encoders with Trojan attacks are as accurate as the classifiers on top of clean SSL image encoders for clean inputs. For example, the accuracy difference between attacked model and the clean model is less than $1\%$.  

\textbf{Effectiveness objective.} The effectiveness objective means that the attack efficacy is very high. In particular, the downstream classifiers built on Trojaned image encoders misclassify the inputs with trigger into the target class with a high probability, e.g., $>99\%$. 

\textbf{Efficiency and stability objective.} The efficiency and stability objective means that attackers can stability achieve the utility and effectiveness goals with as few target samples as possible. This is because identifying target-class samples from large-scale unlabeled training data is expensive. Our attack achieves this goal with only one target sample. 

\subsection{Attacker's Knowledge and Capabilities} We consider that attackers have visibility on a public unlabeled training dataset such that they can identify one target sample for the following attack. Then, attackers pick up an SSL image encoder, add a trigger on the augmentations of training data and optimize for the accuracy and attack efficacy together.  At last, the attackers generate the Trojaned image encoder that achieves attack objectives of utility, effectiveness, efficiency, and stability. Compared to BadEncoders~\cite{jia2022badencoder}, our threat model eliminates the reliance on a pre-trained clean image encoder. Also, our threat model removes the steps in SSL-Backdoor~\cite{saha2021backdoor} to identify $0.5\%$ target samples. There are multiple real-world possible attacks aligned with our threat models, e.g., an untrusted service provider injects the Trojan into the image encoders and releases publicly poisoned encoders with downstream classifiers, a malicious party trains the Trojan encoder and shares it into public model platforms, such as GitHub and ModelZoo~\cite{modelzoo}.

\begin{figure}[t!]
  \centering
   \includegraphics[width=0.9\linewidth]{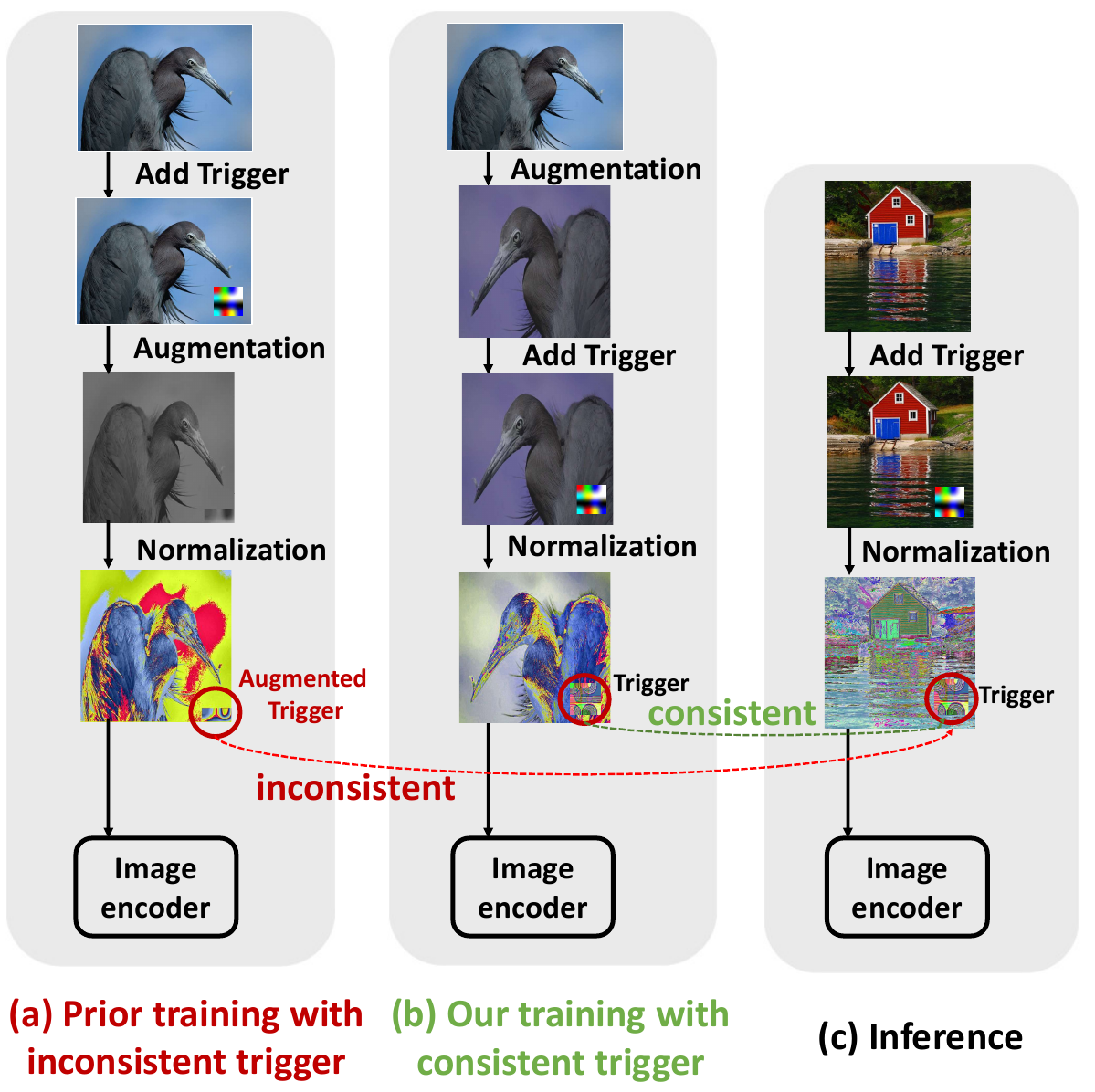}
   \caption{Comparison of trigger poisoning methods. }
   \label{fig:augmentation}
\end{figure}

%%%%%%%%% Our methods
\section{ESTAS Attacks}
\label{sec:methods}
\subsection{Consistent Trigger Poison}
In Figure~\ref{fig:augmentation}(a), we show the attack pipeline of current Trojan attacks in SSL encoders~\cite{saha2021backdoor,liu2022poisonedencoder,jia2022badencoder} that attach the trigger on the original images of the training set, perform the augmentation, e.g., random crops, rotation, color jitter and resize, and normalize the augmentation for the following encoder training. The objective of training an image encoder on the inputs with a trigger is to connect the trigger's representation and target-input representation.  The attack will have a high attack success rate if this connection between the trigger and target input is well-transformed to the inference phase in SSL.  We show the inference phase pipeline in Figure~\ref{fig:augmentation}(c) is asymmetric with image encoding training since the inference phase in SSL has different augmentation that usually only contains resize or has no augmentation~\cite{BYOL:NEURIPS2020}. For this reason, the trigger of previous works used by the image encoder in the training phase, denoted by an augmented trigger in Figure~\ref{fig:augmentation}(a) is inconsistent with the trigger used in the inference, denoted by a trigger in ~\ref{fig:augmentation}(c). Compared to the original trigger used in inference, the augmented trigger may be clipped, pruned, or recolored. This inconsistency significantly limits the transfer of the trigger's feature from the training phase to the inference phase, thus restricting the attack efficacy.

To tackle this problem, we propose an effective method shown in Figure~\ref{fig:augmentation}(b) to design a consistent trigger poison. In particular, we propose to add triggers after augmentation and before normalization in the training, instead of adding triggers before augmentation in previous works. In this way, the trigger used in the image encoder is not affected by data augmentation, thus providing a consistent trigger feature with the inference phase. Our extensive experiments show that our consistent trigger poisoning method promotes building a stronger connection between the trigger and the target class. Specifically, we studied the effects in Tabel~\ref{tab:trigger poison} and it shows that the proposed consistent trigger poisoning method achieves higher $ASR$ than the inconsistent trigger poisoning method.

\subsection{Cascade Optimization Design} 

We propose a cascade optimization method to guide the training of the SSL image encoder that is selected for Trojan insertion. Specifically, our optimization function in ESTAS defined in Equation~\ref{e:loss-estas} consists of two components, i.e., the original loss function of the given SSL method, $L_{S}$, and a cascade loss function for the attack purpose. The cascade loss function includes a local attack loss $L_{L}$ on the SSL encoder and a global attack loss $L_{G}$ on both the encoder and an additional projector, where we consider that the local loss helps extract local features but the global loss on projector's output helps extract additional high-level context information. Our experiments show that combining the local and global loss achieves better feature extraction and Trojan insertion than using one of them. $\lambda_1$, $\lambda_2$ are the hyper-parameters to tune the weights of loss functions, and we use a scheduler to set them from $0$ to $1.0$. Details of the scheduler are shown in Appendix. Our ESTAS method supports different SSL methods, e.g., BYOL~\cite{BYOL:NEURIPS2020} and MoCo v2. We take BYOL~\cite{BYOL:NEURIPS2020} as an example to illustrate the cascade optimization design and leave MoCo v2's details in the Appendix. 

%We use Global Attack Loss combined with Local Attack Loss because on the one hand, triggers are usually small and the output of encoder can keep more details than the projector, so the Local Attack Loss can help the model build stronger trigger-target connection. On the other hand, the Global Attack Loss can help the model consider the overall information and avoid focusing too much on detailed but useless information.

\begin{equation} 
\label{e:loss-estas}
    L_{ESTAS}=L_{S}+\lambda_1 L_{G}+\lambda_2 L_{L}
\end{equation}

\begin{figure}[t!]
  \centering
   \includegraphics[width=\linewidth]{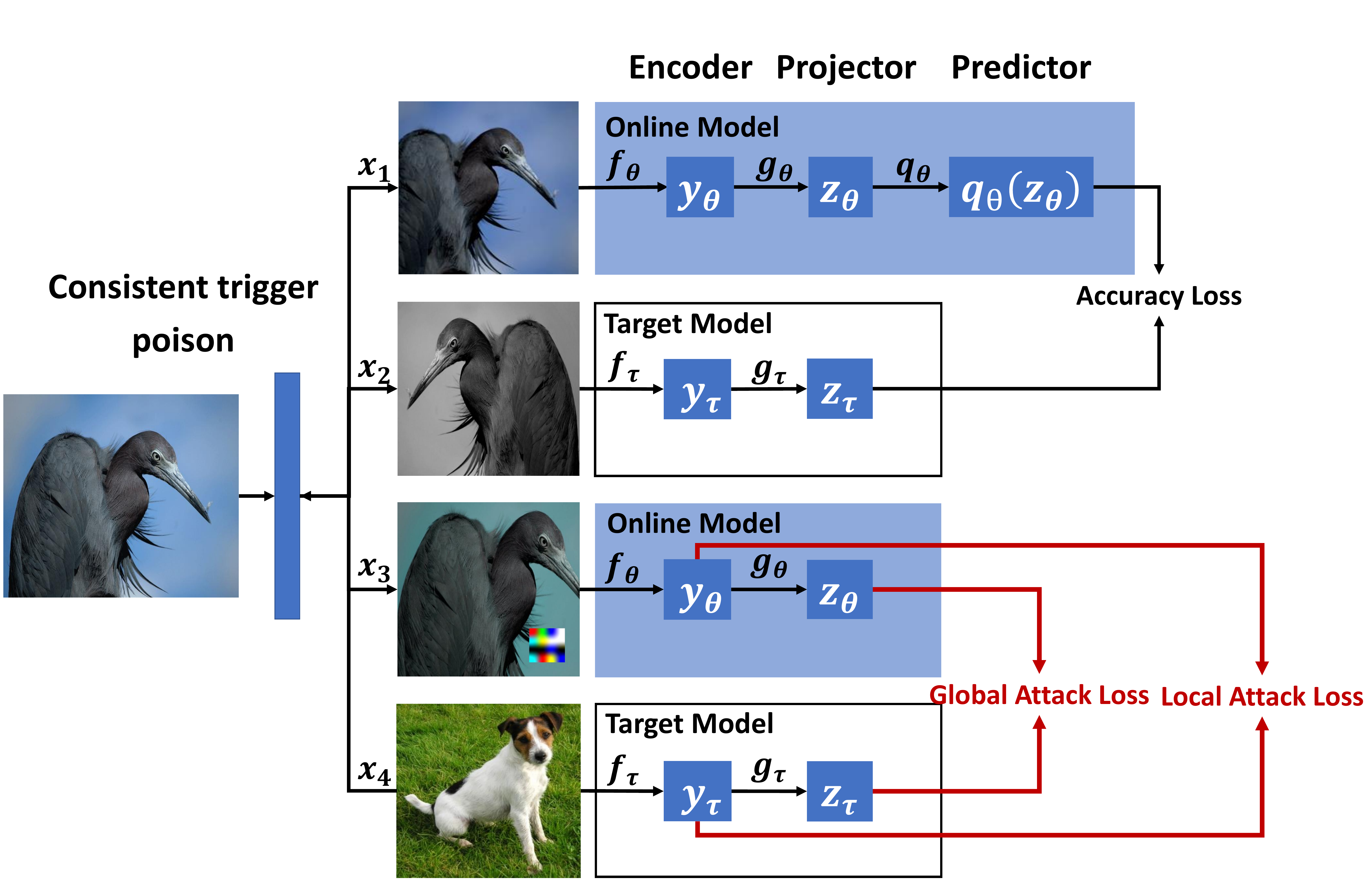}
   \caption{ESTAS cascade optimization with multiple branches.} %Note that the first and third branches have the same online model; the target model is shared by the second and forth branches. }
   \label{fig:byol-loss}
\end{figure}

In Figure~\ref{fig:byol-loss}, we show our attack method with cascade optimization on the BYOL SSL method. For each image $v$ in the training set, we produce three augmented inputs $x_1$, $x_2$, and $x_3$, where only one augmented input, i.e., $x_3$ is attached with a trigger. This process follows the process of our  consistent trigger poison. Our attacks only need one target-class sample. We can generate $x_4$ by applying data augmentation on the target-class sample. We have an online model and a target model. $x_1$ and $x_3$ are used in the online model and $x_2$ and $x_4$ are used in the target model. The target model is updated based on an exponential moving average of the online model weights. The online model has three components, i.e., encoder $f_\theta$, projector $g_\theta$, and predictor $q_\theta$. The target model is comprised of an encoder $f_\tau$ and a projector $g_\tau$. The first branch for augmentation $x_1$ and the third branch for augmentation $x_3$ share the online model, but the third branch does not have a predictor. The second branch for augmentation $x_2$ and the third branch for augmentation $x_4$ share the same target models. In particular, given $x_1$, $x_2$, $x_3$ and $x_4$, we define the outputs of the online encoder, online projector, target encoder, and target projector as  $y_1,y_3=f_\theta(x_1),f_\theta(x_3)$, $z_1,z_3=g_\theta(y_1),g_\theta(y_3)$,  $y_2,y_4=f_\tau(x_2),f_\tau(x_4)$ and  $z_2,z_4=g_\tau(y_2),g_\tau(y_4)$, respectively. Also, we define the predictor in the first branch as $q_\theta$ and its outputs as $q_\theta(z_1)$. Then we normalize $q_\theta(z_1)$, $z_2$, $z_3$ and $z_4$ to $\bar{q}_\theta(z_1)$, $\bar{z}_2$, $\bar{z}_3$ and $\bar{z}_4$.

Based on previous definitions, we describe our loss functions including accuracy loss $L_S$, global attack loss $L_G$, and local attack loss $L_L$ such that one can calculate the loss function $L_{ESTAS}$ in Equation~\ref{e:loss-estas}. In particular, the accuracy loss $L_S$ is defined by a mean squared error between $\bar{q}_\theta(z_1)$ and $\bar{z}_2$ shown in Equation~\ref{e:loss_s} which follows the training loss in BYOL~\cite{BYOL:NEURIPS2020}. $L_S$ is used to achieve our utility goal. 

\begin{equation} 
\label{e:loss_s}
    L_S=||\bar{q}_\theta(z_1)-\bar{z}_2||_2^2=2-2\cdot\frac{<q_\theta\left(z_1\right),z_2>}{\left|\left|q_\theta\left(z_1\right)\right|\right|_2\cdot\left|\left|z_2\right|\right|_2}
\end{equation}

To insert Trojan into the SSL encoder, we promote the model to produce similar representations when the input is a trigger-patched image or a target class's image. For this reason, we define a global attack loss function on the outputs of the projector by a mean squared error between $\bar{z}_3$ and $\bar{z}_4$ shown in Equation~\ref{e:loss_g}.

\begin{equation} 
\label{e:loss_g}
    L_G=||\bar{z}_3-\bar{z}_4||_2^2=2-2\cdot\frac{<z_3,z_4>}{\left|\left|z_3\right|\right|_2\cdot\left|\left|z_4\right|\right|_2}
\end{equation}
Notice here we use the outputs of projector $z_3$ rather than the predictor $q_\theta(z_3)$ in the global loss design. This is because if we use the outputs of the predictor, the encoder will not always build a stable connection between triggers and the target class, and this method may not produce a stable target attack when the encoder is used to build a downstream task. To get a higher $ASR$, we further define a local attack loss on encoder outputs by a mean squared error between $\bar{y}_3$ and $\bar{y}_4$ shown in Equation~\ref{e:loss_l}. Compared to the projector's output, the encoder's output extracts more general and local features. Our experiments and ablation study in the section ~\ref{s:results} shows that our cascade optimization with global attack loss and local attack loss significantly improves the attack efficacy.  
\begin{equation} 
    \label{e:loss_l}
    L_L=||\bar{y}_3-\bar{y}_4||_2^2=2-2\cdot\frac{<y_3,y_4>}{\left|\left|y_3\right|\right|_2\cdot\left|\left|y_4\right|\right|_2}
\end{equation}

%We use Global Attack Loss combined with Local Attack Loss because on the one hand, triggers are usually small and the output of encoder can keep more details than the projector, so the Local Attack Loss can help the model build stronger trigger-target connection. On the other hand, the Global Attack Loss can help the model consider the overall information and avoid focusing too much on detailed but useless information.

%So the model's loss can be defined as, 

%$\lambda_1$, $\lambda_2$ are the hyper-parameters about loss function which are determined according to the importance of each them.

%\subsection{One target sample explanation}

%%%%%%%%% Our methods
\section{Experimental Methodology}
\label{sec:exp}
\subsection{Models and Datasets}
\noindent\textbf{Datasets.}
We perform experiments on CIFAR-10~\cite{cifar10}, CIFAR-100~\cite{cifar100}, and ImageNet-100\cite{imagenet}. CIFAR-10 consists of $50,000$ $32\times32$ color training images and $10,000$ $32\times32$ color testing images in 10 classes; Similar to CIFAR-10, CIFAR-100 has $50,000$ training images and $10,000$ testing images with 100 classes; The ImageNet-100 is a random 100-class subset of ImageNet which is commonly used in self-supervised learning. It has about $127,000$ images in the training set and $5,000$ images in the test set.

\noindent\textbf{SSL Methods.}
We used two representative contrastive learning methods, BYOL~\cite{BYOL:NEURIPS2020} and MoCo v2~\cite{chen2020mocov2}. Consistent with SSL-Backdoor~\cite{saha2021backdoor}, we take  ResNet-18~\cite{resnet} as the encoder for both of our models. For BYOL, the online network and target network have the Projector that is composed of a two-layer multilayer perceptron (MLP) to project 512-dimensional representations to a low-dimensional latent space, i.e., 64-dimensional for CIFAR-10 and CIFAR-100, 128-dimensional for ImageNet-100. In particular, the online network that is used to compute accuracy loss has a predictor composed of two-layer MLP, making the architecture asymmetric; For MoCo v2, the output of the encoder is also 512-dimensional and will be mapped to 128-dimensional by the two-layer MLP Projector.

\noindent\textbf{Linear classifier.}
Consistent with the prior self-supervised backdoor attack, we train a two-layer linear classifier to evaluate the performance of the encoders on downstream supervised tasks. We used a small subset (1\% of the training set) of labeled images of the training set to train the classifier and these images are not used when training the encoder. Then, we get our evaluation metrics through the classifier.

\subsection{Experimental Settings}
\noindent\textbf{Trigger patterns.} For a fair comparison, we adopt the same trigger with with SSL-Backdoor~\cite{saha2021backdoor} which was proposed in Hidden Trigger Backdoor Attacks~\cite{HTBA} firstly. They are square triggers generated by resizing a random $4\times4$ RGB image to the desired patch size using bilinear interpolation. The triggers SSL-Backdoor used are indexed from 10 to 19, and we used the same target-class pair during our experiments. The trigger width is $50\times50$ for ImageNet-100 and $6\times6$ for both CIFAR-100 and CIFAR-10.

\noindent\textbf{Parameter settings.}
For BYOL~\cite{BYOL:NEURIPS2020}, the learning rate is $3e^{-3}$; batch size is $256$ for ImageNet-100, $768$ for CIFAR-10 and CIFAR-100; the target network update rate $\tau$ is $0.99$ with cosine schedule. For MoCo v2~\cite{chen2020mocov2}, learning rate is $3e^{-2}$, batch size is $256$ for ImageNet-100, $768$ for CIFAR-10 and CIFAR-100; the target network update rate $\tau$ is $0.3$. For $\lambda_1$ and $\lambda_2$, they increase linearly from $0$ to $1$ with epoch, and we describe the schedule details in Appendix. 

\subsection{Evaluation Metrics} 
\noindent Clean accuracy ($ACC$): the model accuracy on the clean data. Attack success rate ($ASR$): The $ASR$ is the ratio of successfully predicted images with the trigger to the total number of evaluated images.  The number of unlabeled samples needed to be labeled for poison ($PN$): given a training set size $N$ and a labeled ratio $r$, the attacker needs to identify $PN$ images to attach triggers, $PN=N \times r$. An average number of queries ($AQ$): $AQ$ is the average query number to identify $PN$ images from $N$ unlabeled images. This identification is expensive, especially considering the training set in self-supervised learning is unlabeled and disordered. One needs to check the training set and pick up images belonging to the target class. We show the calculation process of $AQ$ in the appendix.

\section{Results}
\label{s:results}
\subsection{Comparison with prior works}
In Table~\ref{tab:SSLB-imagenet100}, we show that our ESTAS achieves higher attack performance than our baseline attack SSL-Backdoor~\cite{saha2021backdoor} with 10 different trigger-target pairs on ImageNet-100. 
Different from SSL-Backdoor which needs to identify $0.5\%$, i.e., 1270, image samples, our ESTAS only needs to identify one target image. All the three models (Clean Model, SSL-Backdoor, and ESTAS) are trained within 200 epochs with BYOL~\cite{BYOL:NEURIPS2020} or MoCo v2~\cite{chen2020mocov2}. In particular, when the target class is Rottweiler, the clean model achieves  $53.62\%$ and $64.91\%$ accuracy ($ACC-C$) for MoCo v2 and BYOL, respectively. SSL-Backdoor attains $49.90\%$ accuracy ($ACC$) and $9.25\%$ $ASR$ for MoCo v2, $62.46\%$ $ACC$ and $19.52\%$ $ASR$ for BYOL. In contrast, our ESTAS achieved  $53.31\%$ $ACC$ and $85.9\%$ $ASR$ for MoCo v2, $63.09\%$  $ACC$ and $98.58\%$ $ASR$ for BYOL. Therefore,  ESTAS improves $77.7\%$ $ASR$ over SSL-Backdoor. On average, for MoCo v2, SSL-Backdoor achieves $9.3\%$ $ASR$ and $50.1\%$ $ACC$, but ESTAS attains $90.7\%$ $ASR$ and $55.0\%$ $ACC$, thus significantly improving the attack efficacy. For BYOL, SSL-Backdoor achieves $29.1\%$ $ASR$ and $61.6\%$ $ACC$ while ESTAS gets $98.5\%$ $ASR$ and $64.7\%$ $ACC$. Compared with SSL-Backdoor, the performance of ESTAS has been greatly improved, with an average improvement of $75.4\%$ in $ASR$ and $8\%$ in $ACC$. In particular, the $ACC$ of ESTAS is very near to the $ACC-C$ which means our attacks based on the cascade optimization design do not hurt clean accuracy.

\begin{table}[ht!]
  \caption{ESTAS and SSL-Backdoor~\cite{saha2021backdoor} on ImageNet-100}
  \footnotesize
  \setlength{\tabcolsep}{2.5pt}
  \begin{center}
    \begin{NiceTabular}{|c||c||c||c|c||c|c|}[colortbl-like]
    \CodeBefore
        \columncolor{gray!25}{6,7}
    \Body
        \hline
        \multirow{2}{*}{\makecell[c]{Target Class\\(Trigger ID)}} & \multirow{2}{*}{Method} & \multirow{2}{*}{$ACC-C$} & \multicolumn{2}{c|}{SSL-Backdoor} & \multicolumn{2}{c|}{ESTAS}\\ 
        \cline {4-7}
        & & & $ACC$ & $ASR$ & $ACC$ & $ASR$\\\hline\hline
        \multirow{2}{*}{\makecell[c]{Rottweiler\\(10)}} & \multirow{2}{*}{\makecell[c]{MoCo v2\\BYOL}} & \multirow{2}{*}{\makecell[c]{$53.62$\\$64.91$}} & \multirow{2}{*}{\makecell[c]{$49.90$\\$62.46$}} & \multirow{2}{*}{\makecell[c]{$9.25$\\$19.52$}} & \multirow{2}{*}{\makecell[c]{$53.31$\\$63.09$}} & \multirow{2}{*}{\makecell[c]{$85.90$\\$98.58$}}\\
        & & & & & &\\\hline
        \multirow{2}{*}{\makecell[c]{Tabby Cat\\(11)}} & \multirow{2}{*}{\makecell[c]{MoCo v2\\BYOL}} & \multirow{2}{*}{\makecell[c]{$53.62$\\$64.91$}} & \multirow{2}{*}{\makecell[c]{$50.48$\\$62.26$}} & \multirow{2}{*}{\makecell[c]{$29.90$\\$37.76$}} & \multirow{2}{*}{\makecell[c]{$59.98$\\$64.27$}} & \multirow{2}{*}{\makecell[c]{$93.57$\\$98.91$}}\\
        & & & & & &\\\hline
        \multirow{2}{*}{\makecell[c]{Ambulance\\(12)}} & \multirow{2}{*}{\makecell[c]{MoCo v2\\BYOL}} & \multirow{2}{*}{\makecell[c]{$53.62$\\$64.91$}} & \multirow{2}{*}{\makecell[c]{$50.80$\\$60.88$}} & \multirow{2}{*}{\makecell[c]{$2.08$\\$18.51$}} & \multirow{2}{*}{\makecell[c]{$55.15$\\$65.38$}} & \multirow{2}{*}{\makecell[c]{$92.03$\\$98.31$}}\\
        & & & & & &\\\hline
        \multirow{2}{*}{\makecell[c]{Pickup Truck\\(13)}} & \multirow{2}{*}{\makecell[c]{MoCo v2\\BYOL}} & \multirow{2}{*}{\makecell[c]{$53.62$\\$64.91$}} & \multirow{2}{*}{\makecell[c]{$50.58$\\$61.28$}} & \multirow{2}{*}{\makecell[c]{$1.94$\\$7.64$}} & \multirow{2}{*}{\makecell[c]{$51.93$\\$67.45$}} & \multirow{2}{*}{\makecell[c]{$91.83$\\$98.05$}}\\
        & & & & & &\\\hline
        \multirow{2}{*}{\makecell[c]{Laptop\\(14)}} & \multirow{2}{*}{\makecell[c]{MoCo v2\\BYOL}} & \multirow{2}{*}{\makecell[c]{$53.62$\\$64.91$}} & \multirow{2}{*}{\makecell[c]{$49.78$\\$61.64$}} & \multirow{2}{*}{\makecell[c]{$10.61$\\$36.83$}} & \multirow{2}{*}{\makecell[c]{$54.03$\\$65.82$}} & \multirow{2}{*}{\makecell[c]{$92.37$\\$96.06$}}\\
        & & & & & &\\\hline
        \multirow{2}{*}{\makecell[c]{Goose\\(15)}} & \multirow{2}{*}{\makecell[c]{MoCo v2\\BYOL}} & \multirow{2}{*}{\makecell[c]{$53.62$\\$64.91$}} & \multirow{2}{*}{\makecell[c]{$50.70$\\$62.04$}} & \multirow{2}{*}{\makecell[c]{$7.11$\\$53.23$}} & \multirow{2}{*}{\makecell[c]{$57.11$\\$63.73$}} & \multirow{2}{*}{\makecell[c]{$86.90$\\$98.80$}}\\
        & & & & & &\\\hline
        \multirow{2}{*}{\makecell[c]{Pirate Ship\\(16)}} & \multirow{2}{*}{\makecell[c]{MoCo v2\\BYOL}} & \multirow{2}{*}{\makecell[c]{$53.62$\\$64.91$}} & \multirow{2}{*}{\makecell[c]{$49.68$\\$61.72$}} & \multirow{2}{*}{\makecell[c]{$9.41$\\$21.11$}} & \multirow{2}{*}{\makecell[c]{$53.91$\\$66.18$}} & \multirow{2}{*}{\makecell[c]{$91.62$\\$98.81$}}\\
        & & & & & &\\\hline
        \multirow{2}{*}{\makecell[c]{Gas Mask\\(17)}} & \multirow{2}{*}{\makecell[c]{MoCo v2\\BYOL}} & \multirow{2}{*}{\makecell[c]{$53.62$\\$64.91$}} & \multirow{2}{*}{\makecell[c]{$49.60$\\$60.60$}} & \multirow{2}{*}{\makecell[c]{4.75\\74.38}} & \multirow{2}{*}{\makecell[c]{$56.70$\\$67.29$}} & \multirow{2}{*}{\makecell[c]{$91.65$\\$97.69$}}\\
        & & & & & &\\\hline
        \multirow{2}{*}{\makecell[c]{Vacuum Cleaner\\(18)}} & \multirow{2}{*}{\makecell[c]{MoCo v2\\BYOL}} & \multirow{2}{*}{\makecell[c]{$53.62$\\$64.91$}} & \multirow{2}{*}{\makecell[c]{$49.82$\\$62.36$}} & \multirow{2}{*}{\makecell[c]{$4.91$\\$5.84$}} & \multirow{2}{*}{\makecell[c]{$54.82$\\$64.33$}} & \multirow{2}{*}{\makecell[c]{$83.72$\\$98.05$}}\\
        & & & & & &\\\hline
        \multirow{2}{*}{\makecell[c]{American Lobster\\(19)}} & \multirow{2}{*}{\makecell[c]{MoCo v2\\BYOL}} & \multirow{2}{*}{\makecell[c]{$53.62$\\$64.91$}} & \multirow{2}{*}{\makecell[c]{$50.02$\\$60.98$}} & \multirow{2}{*}{\makecell[c]{$13.19$\\$16.57$}} & \multirow{2}{*}{\makecell[c]{$52.93$\\$63.02$}} & \multirow{2}{*}{\makecell[c]{$97.20$\\$99.99$}}\\
        & & & & & &\\\hline
        \multirow{2}{*}{Average} & \multirow{2}{*}{\makecell[c]{MoCo v2\\BYOL}} & \multirow{2}{*}{\makecell[c]{$53.6$\\$64.91$}} & \multirow{2}{*}{\makecell[c]{$50.1$\\$61.6$}} & \multirow{2}{*}{\makecell[c]{$9.3$\\$29.1$}} & \multirow{2}{*}{\makecell[c]{$55.0$\\$64.7$}} & \multirow{2}{*}{\makecell[c]{$90.7$\\$98.5$}}\\
        & & & & & &\\\hline
    \end{NiceTabular}%
  \label{tab:SSLB-imagenet100}%
  \end{center}
\end{table}%

In Table~\ref{tab:other baseline}, we compare ESTAS with other related Trojan attacks in SSL image encoders. They adopt $0.5\sim1\%$  poisoning ratio on training datasets with BYOL~\cite{BYOL:NEURIPS2020}. Our ESTAS achieves superior performance, e.g., $ACC$, $ASR$, attack cost, over previous works.  In particular, ESTAS gets $87.3\%$, $67.3\%$ and $62.7\%$ $ACC$ on CIFAR-10, cifar-100, and ImageNet-100, respectively, achieving $6.8\%$, $12.1\%$ and $8.7\%$ improvement over the other three attacks on average. This is because ESTAS split the training loss into accuracy loss and attack loss and only the clean images are involved to compute the accuracy loss, which eliminates the impact of the poisoned images on clean accuracy.

Also, ESTAS achieves $100\%$, $98.9\%$ and $98.5\%$ $ASR$ on CIFAR-10, CIFAR-100, and ImageNet-100, respectively, which increases $46.2\%$, $63.3\%$ and $60.7\%$ $ASR$ over 
other three attacks on average. This is because, in SSL-Backdoor~\cite{saha2021backdoor}, Poisoned-Encoder and CTRL~\cite{li2022demystifying}, images with the trigger only account for $1\%$ of the whole dataset, meaning that it is hard for the model to learn the connection between the trigger and target-class images. And, the trigger is only attached to the target class's images so the model will only learn that the trigger is a feature of the target class, but will not build an instant connection between the trigger and the target. In contrast, our ESTAS uses consistent trigger poison so that the model  learns a better trigger pattern.

%\textbf{ESTAS has the lowest cost.}
The average number of queries ($AQ$) of ESTAS is much smaller than other attacking methods, which means the ESTAS is more efficient to perform Trojan attacks. Especially for ImageNet-100, SSL-Backdoor~\cite{saha2021backdoor} and Poisoned-Encoder~\cite{liu2022poisonedencoder} need to query 126,901 images on average to poison $1\%$ image, which is very expensive. Such a large query number means the attacker needs to look through, comparable to annotating and labeling, almost the whole training set to identify the images to be poisoned. In contrast, our attacks significantly reduce the average number, thus reducing the attacking cost.
\begin{table}[ht!]
  \caption{Comparison of ESTAS and other attack methods}
  \footnotesize
  \setlength{\tabcolsep}{5pt}
  \begin{center}
    \begin{NiceTabular}{|c||c||c||c||c||c|}[colortbl-like]
    \CodeBefore
        \rowcolor{gray!25}{14,15,16}
    \Body
        \hline
        Attack Method & Dataset & $ACC$ & $ASR$ & $PN$ & $AQ$ \\
        \hline
        \hline
        \multirow{3}{*}{Clean Model} & CIFAR-10 & 86.9 & 3.1 & 0 & 0 \\
        & CIFAR-100 & $67.4$ & $0.8$ & $0$ & $0$ \\
        & ImageNet-100 & $64.9$ & $0.1$ & $0$ & $0$ \\
        \hline
        \multirow{3}{*}{\makecell{SSL\\-Backdoor}} & CIFAR-10 & $82.3$ & $47.2$ & $500$ & $4912$ \\
        & CIFAR-100 & $50.4$ & $26.9$ & $500$ & $49901$ \\
        & ImageNet-100 & $60.6$ & $33.2$ & $1270$ & $126901$ \\
        \hline
        \multirow{3}{*}{\makecell{Poisoned\\-Encoder}} & CIFAR-10 & $81.7$ & $50.7$ & $500$ & $4912$ \\
        & CIFAR-100 & $50.9$ & $32.5$ & $500$ & $49901$ \\
        & ImageNet-100 & $48.1$ & $38.1$ & $1270$ & $126901$ \\
        \hline
        \multirow{3}{*}{CTRL} & CIFAR-10 & $77.6$ & $63.5$ & $500$ & $4912$ \\
        & CIFAR-100 & $54.1$ & $47.3$ & $500$ & $49901$ \\
        & ImageNet-100 & $53.3$ & $42.1$ & $1270$ & $126901$ \\
        \hline
        \multirow{3}{*}{ESTAS} & CIFAR-10 & $87.3$ & $100$ & $1$ & $10$ \\
        & CIFAR-100 & $67.3$ & $98.9$ & $1$ & $100$ \\
        & ImageNet-100 & $62.7$ & $98.5$ & $1$ & $100$ \\
        \hline
    \end{NiceTabular}%
  \label{tab:other baseline}%
  \end{center}
\end{table}%e 2

In Table~\ref{tab:BadEncoder}, we compare ESTAS with another related work BadEncoder~\cite{jia2022badencoder}. The comparison shows that our attack is more stable when using one target-class sample.  We select Tabby Cat as our target class and randomly select 10 different images belonging to the target class as the poisoned images. The poisoned input is the index of the poison image used to attack, and the mistaken target means the model has a high $ASR$ on this class but it is not our target class. By trying 10 different images, BadEncoder only attacks successfully on 3 images but will attack other class which is similar to the target class by mistake, i.e., House Cat and Ally Cat. However, ESTAS attacks successfully on all images with higher $ASR$. On average, The $ASR$ of ESTAS is $77\%$ higher than that of BadEncoder. We believe this is because Alley Cat and House Cat are similar to Tabby Cat in feature space so the pre-trained encoder cannot separate them well in representation space. Therefore, when we input a poisoned image, there is a probability to attack the wrong class. Null in Mistaken Target column means the attack is successful.

\begin{table}[ht!]
  \caption{Results of ESTAS and BadEncoder on ImageNet-100}
  \footnotesize
  \setlength{\tabcolsep}{4pt}
  \begin{center}
    \begin{NiceTabular}{|c||c|c|c||c|c|c|}[colortbl-like]
    \CodeBefore
        \columncolor{gray!25}{5,6,7}
    \Body
        \hline
        \multirow{3}{*}{\makecell{Poisoned\\Input}} & \multicolumn{3}{c|}{BadEncoder} & \multicolumn{3}{c|}{ESTAS}\\
        \cline{2-7}
        & \multirow{2}{*}{$ACC$} & \multirow{2}{*}{$ASR$} & \multirow{2}{*}{\makecell{Mistaken\\Target}} & \multirow{2}{*}{$ACC$} & \multirow{2}{*}{$ASR$} & \multirow{2}{*}{\makecell{Mistaken\\Target}}\\
        & & & & & &\\
        \hline\hline
        Tabby Cat 0 & $66.13$ & $0.42$ & House Cat & $67.61$ & $99.42$ & Null\\
        \hline
        Tabby Cat 1 & $62.48$ & $0.26$ & House Cat & $64.93$ & $99.08$ & Null\\
        \hline
        Tabby Cat 2 & $63.62$ & $1.03$ & Alley Cat & $65.01$ & $99.21$ & Null\\
        \hline
        Tabby Cat 3 & $59.09$ & $0.91$ & Alley Cat & $67.99$ & $98.96$ & Null\\
        \hline
        Tabby Cat 4 & $59.55$ & $46.20$ & Toy Terrier & $67.87$ & $97.32$ & Null\\
        \hline
        Tabby Cat 5 & $61.91$ & $0.05$ & Alley Cat & $64.35$ & $98.95$ & Null\\
        \hline
        Tabby Cat 6 & $61.34$ & $75.31$ & Null & $63.91$ & $94.10$ & Null\\
        \hline
        Tabby Cat 7 & $57.55$ & $3.22$ & House Cat & $64.87$ & $92.72$ & Null\\
        \hline
        Tabby Cat 8 & $60.91$ & $77.92$ & Null & $66.35$ & $99.90$ & Null\\
        \hline
        Tabby Cat 9 & $61.34$ & $1.37$ & Alley Cat & $67.91$ & $95.42$ & Null\\
        \hline
        Average & $61.4$ & $20.5$ & & $66.1$ & $97.5$ & \\
        \hline
    \end{NiceTabular}%
  \label{tab:BadEncoder}%
  \end{center}
\end{table}%

\subsection{Ablation study}
\noindent\textbf{Consistent trigger poison evaluation.} In Table~\ref{tab:trigger poison}, we compare the attack effectiveness of ESTAS using consistent trigger poison and ESTAS using inconsistent trigger poison on CIFAR-10, CIFAR-100, and ImageNet-100. On CIFAR-10, ESTAS achieves $100\%$ $ASR$ and $87.3\%$ $ACC$ with consistent trigger poison, but only achieves $90.4\%$ $ASR$ and $87.1\%$ $ACC$ with inconsistent trigger poison. On CIFAR-100, ESTAS obtains $98.9\%$ $ASR$ with $67.3\%$ $ACC$ using consistent trigger poison, but only achieves $85.9\%$ $ASR$ and $68.1\%$ $ACC$ with inconsistent trigger poison. On ImageNet-100, ESTAS gets $98.5\%$ $ASR$ and $62.7\%$ $ASR$ with consistent trigger poison, but only obtains $91.2\%$ $ASR$ with $61.9\%$ $ACC$ without consistent trigger poison. This result shows that using Consistent trigger poison can significantly improve $ASR$ without any impact on $ACC$.

\begin{table}[ht!]
  \caption{Performance of ESTAS under consistent trigger poison and inconsistent trigger poison}
  \footnotesize
  \setlength{\tabcolsep}{2.5pt}
  \begin{center}
    \begin{NiceTabular}{|c||>{\centering\arraybackslash}p{1.4cm}|>{\centering\arraybackslash}p{1.4cm}||>{\centering\arraybackslash}p{1.4cm}|>{\centering\arraybackslash}p{1.4cm}|}[colortbl-like]
    \CodeBefore
        \columncolor{gray!25}{5,4}
    \Body
        \hline
        \multirow{2}{*}{Dataset} & \multicolumn{2}{c|}{Inconsistent trigger poison} & \multicolumn{2}{c|}{Consistent trigger poison}\\
        \cline{2-5}
        & $ACC$ & $ASR$ & $ACC$ & $ASR$\\
        \hline\hline
        CIFAR-10 & $87.1$ & $90.4$ & $87.3$ & $100$\\
        \hline
        CIFAR-100 & $68.1$ & $85.9$ & $67.3$ & $98.9$\\
        \hline
        ImageNet-100 & $61.9$ & $91.2$ & $62.7$ & $98.5$\\
        \hline
    \end{NiceTabular}%
  \label{tab:trigger poison}%
  \end{center}
  \vspace{-0.2in}
\end{table}%

\noindent\textbf{Attack loss evaluation.} 
In Table~\ref{tab:loss}, we conduct an ablation study to find out the best design of attack loss. We randomly select 5 trigger-target pairs on CIFAR-10. When using the outputs of the predictor to compute the attack loss $L_A=||\bar{q}_\theta(z_3)-\bar{z}_4||_2^2$.
%\begin{equation}
%\label{e:loss_A1}
%    L_A=||\bar{q}_\theta(z_3)-\bar{z}_4||_2^2=2-2\cdot\frac{<q_\theta\left(z_3\right),z_4>}{\left|\left|q_\theta\left(z_3\right)\right|\right|_4\cdot\left|\left|z_4\right|\right|_4}
%\end{equation}
The $ASR$ is low and unstable as the result shows. This is because the encoder does not build a connection between the trigger and the target.
When only using $L_G$, the $ASR$ is more stable but not very high, $70.0\%$ on average. This is because the projector will filter detailed information about the trigger. And only using $L_L$, the $ASR$ can be higher but less stable. We believe this is because it focuses on the details but omits the overall information. 

The performance is best when we use both global attack loss and local attack loss: the $ASR$ is more stable and the average $ASR$ is $99.9\%$. Therefore, the combination of global attack loss and local attack loss is the key to the attack loss design.

\begin{table}[ht!]
  \caption{Ablation study on attack loss design}
  \footnotesize
  \setlength{\tabcolsep}{2.5pt}
  \begin{center}
    \begin{NiceTabular}{|c||c|c||c|c||c|c||c|c|}[colortbl-like]
    \CodeBefore
        \columncolor{gray!25}{8,9}
    \Body
        \hline
        \multirow{2}{*}{\makecell{Target Class\\(Trigger ID)}} & \multicolumn{2}{c|}{Use Predictor} & \multicolumn{2}{c|}{Global} & \multicolumn{2}{c|}{Local} & \multicolumn{2}{c|}{Global+Local}\\
        \cline{2-9}
        & $ACC$ & $ASR$ & $ACC$ & $ASR$ & $ACC$ & $ASR$ & $ACC$ & $ASR$\\ 
        \hline\hline
        Airplane(10) & $86.6$ & $3.4$ & $87.5$ & $88.8$ & $87.3$ & $97.2$ & $88.3$ & $99.9$\\
        \hline
        Bird(12) & $87.1$ & $5.9$ & $87.6$ & $91.5$ & $87.8$ & $4.9$ & $87.0$ & $100$\\
        \hline
        Deer(14) & $86.9$ & $61.1$ & $86.3$ & $83.2$ & $86.1$ & $91.2$ & $87.4$ & $99.8$\\
        \hline
        Frog(16) & $87.5$ & $1.3$ & $87.9$ & $78.2$ & $87.6$ & $43.9$ & $87.0$ & $99.9$\\
        \hline
        Ship(18) & $87.1$ & $5.0$ & $88.2$ & $82.3$ & $86.9$ & $98.8$ & $87.9$ & $100$\\
        \hline
        Average & $87$ & $15.3$ & $87.5$ & $70.0$ & $87.1$ & $67.2$ & $87.7$& $99.9$\\
        \hline
    \end{NiceTabular}%
  \label{tab:loss}%
  \end{center}
\end{table}%

\subsection{Potential Defense} 

We distill the encoder poisoned by ESTAS using CompRess~\cite{compress,9360315} on a subset of ImageNet-100. The result shows that using distillation can defend against ESTAS to a certain extent, but the cost is that the $ACC$ will be reduced. For example, when using $10\%$ of ImageNet-100 to distillate the model, $ASR$ drops from $98.5\%$ to $77.4\%$ while the $ACC$ drops from $64.7\%$ to $51.8\%$.

\begin{table}[ht!]
  \caption{Defense with compress distillation}
  \small
  \setlength{\tabcolsep}{5pt}
  \begin{center}
    \begin{NiceTabular}{|c||c||c|}
        \hline
        Method & $ACC$ & $ASR$\\
        \hline\hline
        ESTAS & $64.7$ & $98.5$\\
        \hline
        Defense $25\%$ & $60.2$ & $63.0$ \\
        \hline
        Defense $10\%$ & $51.8$ & $77.4$ \\
        \hline
        Defense $5\%$ & $43.9$ & $83.1$\\
        \hline
    \end{NiceTabular}%
  \label{tab:defense}%
  \end{center}
  \vspace{-0.2in}
\end{table}%

\section{Conclusion}
In this work, we present ESTAS that enables effective and stable Trojan attacks in SSL image encoders with even one target-class image sample. ESTAS depends on two key components, i.e., consistent trigger poison and cascade optimization, to achieve superior performance. Our comprehensive experiments show that  ESTAS attains $>30\%$ $ASR$ increase and $>8.3\%$ accuracy improvement on average over prior works. And ESTAS stably achieves $>99\%$ $ASR$ with one target-class sample on multiple tasks.

\section{Appendix}
%%%%%%%%% BODY TEXT
\subsection{ESTAS on MoCo v2}
\label{sec:MoCo v2}
\begin{figure}[t!]
  \centering
   \includegraphics[width=\linewidth]{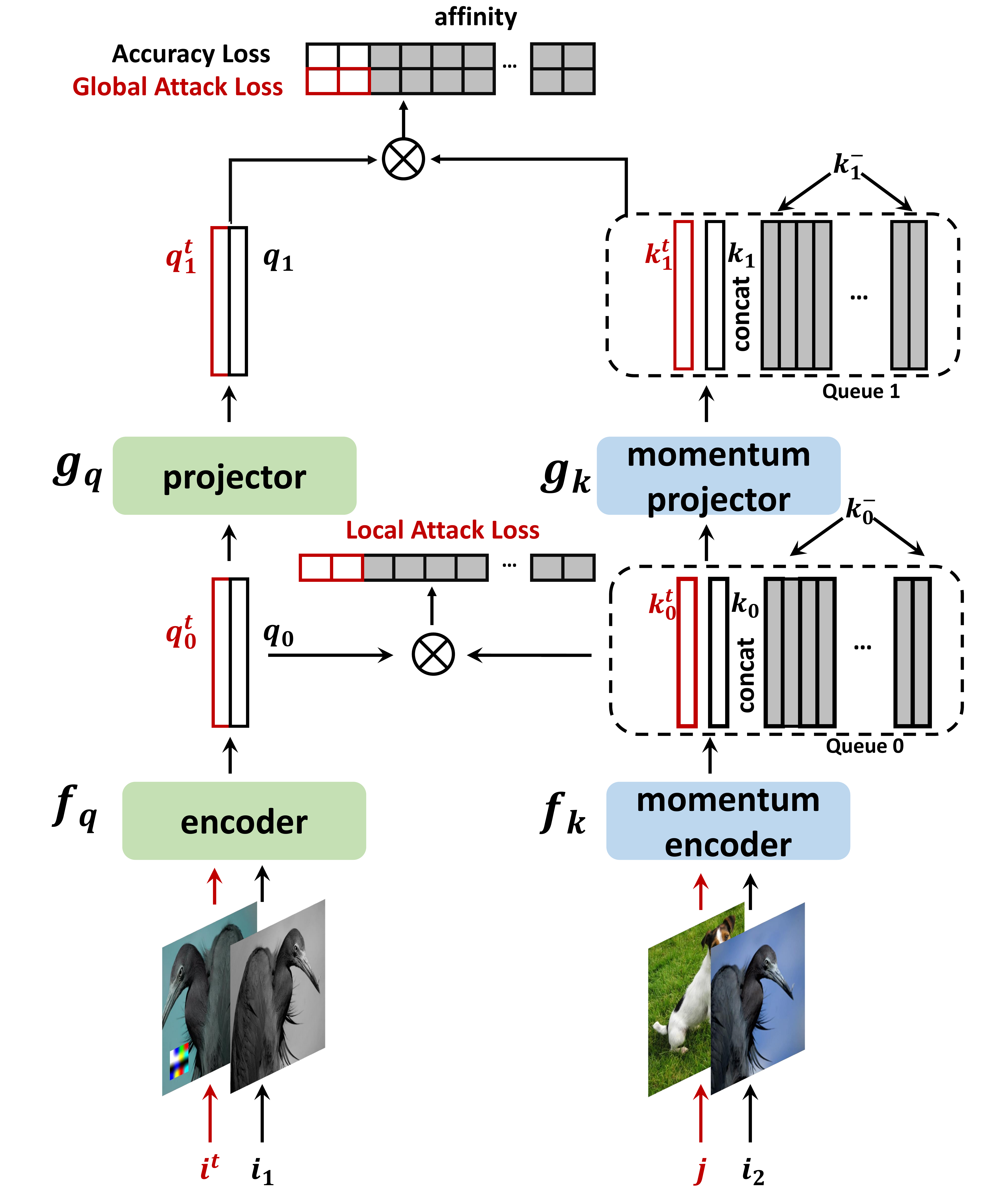}
   \caption{ESTAS with MoCo v2.} %Note that the first and third branches have the same online model; the target model is shared by the second and forth branches. }
   \label{fig:moco-loss}
\end{figure}

In Figure~\ref{fig:moco-loss}, we show our attack method on the MoCo v2 SSL method. First, given any unlabeled image, we use data augmentation in MoCo V2 to generate three augmentations, i.e, $i$, $i_1$, and $i_2$. Only the augmentation $i$ is attached trigger to obtain the poisoned image $i^t$ by our proposed consistent trigger poison method. Our ESTAS only needs one target-class sample and we denote its argumentation as $j$.  We load $i^t$ and $i_1$ into encoder $f_q$ to derive representations $q_0^t$ and $q_0$, respectively, and put $j$ and $i_2$ into momentum encoder $f_k$ to obtain representations $k_0^t$ and $k_0$. We also use Equation~\ref{e:q_0} and Equation~\ref{e:k_0} to show the computing process of encoded representations.    
\begin{equation} 
\label{e:q_0}
    q_0^t,q_0=f_q(i^t),f_q(i_1)
\end{equation}
\begin{equation} 
\label{e:k_0}
    k_0^t,k_0=f_k(j),f_k(i_2)
\end{equation}
As the Equation~\ref{e:q_1} and Equation~\ref{e:k_1} show, the projector $g_q$ and momentum projector $g_k$ will project the generated representations $q_0^t$, $q_0$, $k_0^t$ and $k_0$ with a high dimensional space to $q_1^t$, $q_1$, $k_1^t$ and $k_1$ with a lower dimensional space.

\begin{equation} 
\label{e:q_1}
    q_1^t,q_1=g_q(q_0^t),g_q(q_0)
\end{equation}
\begin{equation} 
\label{e:k_1}
    k_1^t,k_1=g_k(k_0^t),g_k(k_0)
\end{equation}

\textbf{Cascade optimization design.}
We first define the accuracy loss $L_S$ by following MoCo v2~\cite{chen2020mocov2} as Equation~\ref{e:accuracy loss} shows. Here ${k_1^-}$ are the representations of negative (dissimilar) key samples stored in a queue, e.g., Queue 1 in Figure~\ref{fig:moco-loss}. $\tau$ is the temperature hyper-parameter, and the default $\tau=0.2$ is used, as in ~\cite{chen2020mocov2}. For each query, $q_1$ and $k_1$ form a positive pair because they are data-augmented versions of the same image. In contrast, $q_1$ and ${k_1^-}$ form negative pairs since they come from different-class images. After each query, $k_0$ and $k_1$ will be appended to Queue 0 and Queue 1 respectively, acting as negative samples in other queries. 
\begin{equation} 
\label{e:accuracy loss}
    L_S=-log\frac{exp\left(q_1\cdot k_1/\tau\right)}{exp\left(q_1\cdot k_1/\tau\right)+\sum_{k_1^-}exp\left(q_1\cdot k_1^-/\tau\right)}
\end{equation}

\begin{figure*}[t!]
  \centering
   \includegraphics[width=0.8\linewidth]{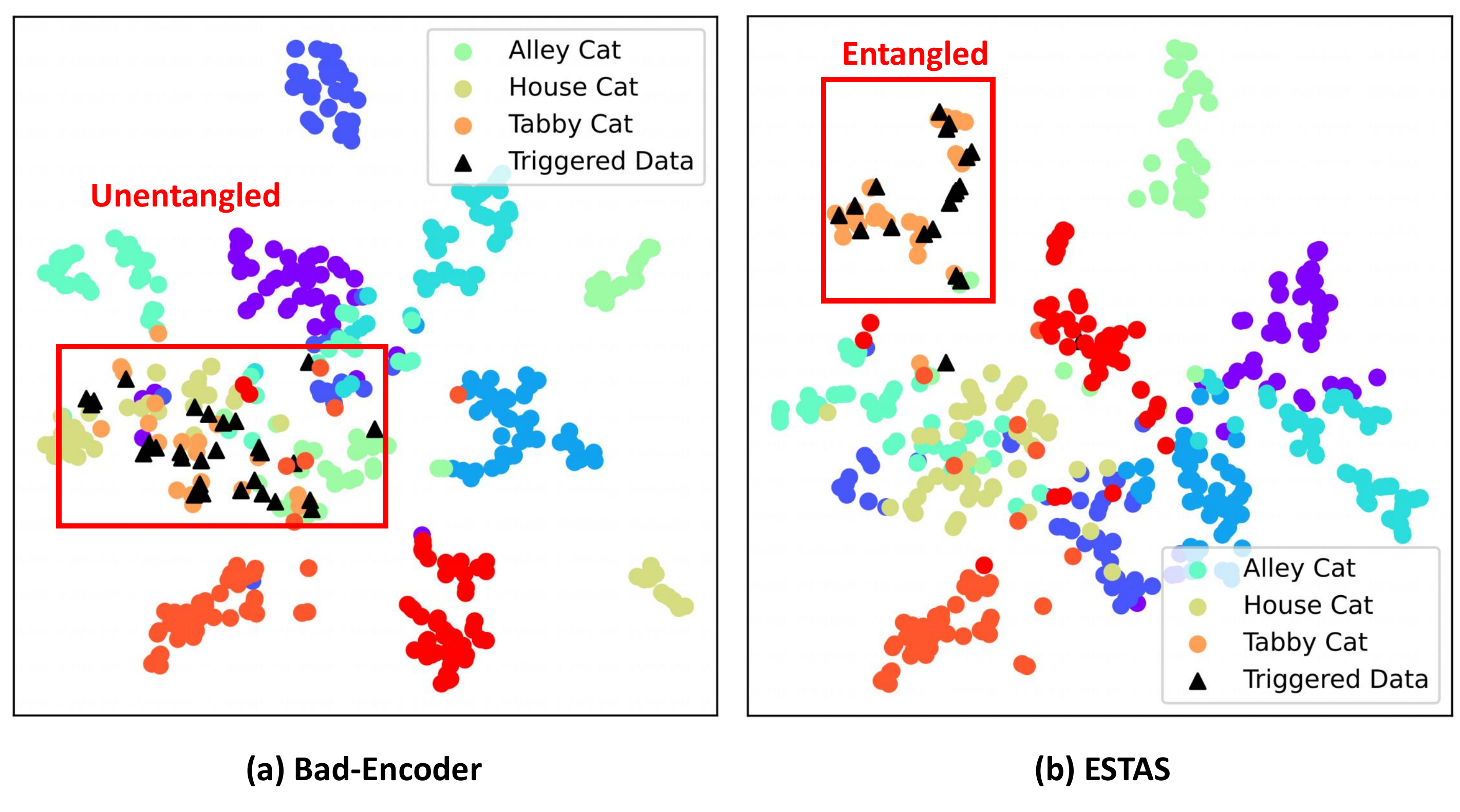}
   \caption{t-SNE visualization of features in Bad-Encoder.} %Note that the first and third branches have the same online model; the target model is shared by the second and fourth branches. }
   \label{fig:tSNE}
\end{figure*}
We formulate our attack in SSL encoder as one optimization problem, i.e., maximizing the similarity between representations $q_0^t$ and $k_0^t$, $q_1^t$ and $k_1^t$. For this reason, we define local attack loss $L_L$ and global attack loss $L_G$ in Equation~\ref{e:local attack loss} and Equation~\ref{e:global attack loss}, respectively. $q_0^t$ and $q_1^t$ are representations of the image that is poisoned, $k_0^t$ and $k_1^t$ are representations of the image that is a target class, and $k_0^-$ and $k_1^-$ are representations of other images which means $q_0^t$ and $k_0^t$, $q_1^t$ and $k_1^t$ form positive pairs whereas $q_0^t$ and $k_0^-$, $q_1^t$ and $k_1^-$ form negative pairs. We use both local attack loss and global attack loss because we consider the local loss helps extract local features but the global loss on the projector’s output helps extract additional high-level context information.
\begin{equation} 
\label{e:local attack loss}
    L_L=-log\frac{exp\left(q_0^t\cdot k_0^t/\tau\right)}{exp\left(q_0^t\cdot k_0^t/\tau\right)+\sum_{k_0^-}exp\left(q_0^t\cdot k_0^-/\tau\right)}
\end{equation}
\begin{equation} 
\label{e:global attack loss}
    L_G=-log\frac{exp\left(q_1^t\cdot k_1^t/\tau\right)}{exp\left(q_1^t\cdot k_1^t/\tau\right)+\sum_{k_1^-}exp\left(q_1^t\cdot k_1^-/\tau\right)}
\end{equation}

\subsection{Average number of queries}
\label{sec:AQ}
Given a ratio $r$ and an unlabeled training set with $N$ images including $N_t$ target-class images, the attacker needs to identify $PN=N\times r$ target-class images to poison. To identify one target-class image from the unlabeled and disordered dataset, the attacker must go through the whole dataset. We define the process of determining whether an image belongs to the target class as a single query. The average number of queries ($AQ$) to get $PN$ target-class samples is the mathematic expectation of the number of queries.

First, we define $P(n)$ in Equation~\ref{e:P} to describe the probability that at least $n$ queries are needed to get $PN$ target-class images.
\begin{equation} 
    \label{e:P}
        P(n)=\frac{\binom{N_t}{PN}\cdot\binom{N-N_t}{n-PN}}{\binom{N}{n}}\cdot\frac{PN}{n}
\end{equation}

The first term of Equation~\ref{e:P}, i.e., $\frac{\binom{N_t}{PN}\cdot\binom{N-N_t}{n-PN}}{\binom{N}{n}}$ denotes the probability of getting $PN$ target-class images in $n$ queries and it can be understood as follows: $\binom{N}{n}$ denotes the number of methods to draw $n$ out of $N$ images; $\binom{N_t}{PN}$ indicates the number of methods to draw $N_t$ out of $PN$ target-class images and $\binom{N-N_t}{n-PN}$ indicates the number of methods to draw $N-N_t$ out of $n-PN$ non-target class images. The second term, i.e., $\frac{PN}{n}$, is to ensure that the last image of $n$ belongs to the target class. Otherwise, it will no longer need $n$ queries to get $N_t$ target-class images.

So, we can define $AQ$ using the following Equation~\ref{e:AQ} and Equation~\ref{e:K}, where $K$ in Equation~\ref{e:AQ} and Equation~\ref{e:K} means the worst-case query number. The worst case means that all $N_t$ target-class images are located in the last $N_t$ images.

\begin{equation} 
    \label{e:AQ}
        AQ=E(n)=\sum_{n=1}^KnP(n)=\sum_{n=1}^K\frac{\binom{N_t}{PN}\cdot\binom{N-N_t}{n-PN}}{\binom{N}{n}}\cdot PN
\end{equation}

\begin{equation} 
    \label{e:K}
        K=N-PN+N_t
\end{equation}

\subsection{Stability and Attack Efficacy Analysis on Bad-Encoder and our ESTAS}
\label{sec:tSNE}

To compare the stability and attack efficacy of our baseline Bad-Encoder and our ESTAS, in Figure~\ref{fig:tSNE}, we use t-SNE~\cite{tSNE} to visualize their representations of poisoned inputs and clean inputs in the test set. Figure~\ref{fig:tSNE}(a) shows the distribution of representations when attacking a pre-trained clean model with Bad-Encoder\cite{jia2022badencoder}. The representations of the poisoned images are not only entangled with the target-class representations, i.e., Tabby Cat, but also entangled with the House Cat and Alley Cat that have similar representations with Tabby Cat in feature space. This is because the clean pre-trained encoder will output similar representations to these categories. This is one of the reasons why Bad-Encoder can occasionally achieve a high $ASR$ but cannot achieve a stable attack, e.g., it may classify the poisoned image into a non-target class that is similar to the target class. In contrast, when using ESTAS, as Figure~\ref{fig:tSNE}(b) shows, the encoder generates entangled representations for poisoned images and target-class images. And their representations have a large distance from other-class features that are similar in the Bad-Encoder.  One of the reasons is that our ESTAS trains an encoder from scratch, eliminating the reliance on a pre-trained model that is required by our baseline Bad-Encoder for attacks.

\subsection{Scheduler of hyper-parameters $\lambda_1$ and $\lambda_2$ for multiple-object optimization}
\label{sec:loss}

\begin{equation} 
\label{e:loss-estas-ap}
    L_{ESTAS}=L_{S}+\lambda_1 L_{G}+\lambda_2 L_{L}
\end{equation}

\begin{equation} 
\label{e:lambda}
    \lambda_1,\lambda_2=\lambda_{10} \cdot \frac{e}{E},\lambda_{20} \cdot \frac{e}{E}
\end{equation}

Our ESTAS converts the Trojan attack in the SSL image encoder as a multiple-object optimization problem, i.e., optimizing model accuracy, global attack loss, and local attack loss shown in Equation~\ref{e:loss-estas-ap}. For this reason, we have two hyper-parameters  $\lambda_1$ and $\lambda_2$. We use a linear scheduler to enable stable training.  The $\lambda_1$ and $\lambda_2$ will change according to current training epochs $e$ and total training epochs $E$ as Equation~\ref{e:lambda} shows. In Equation~\ref{e:lambda}, $\lambda_{10}$ and $\lambda_{20}$ are the initial values of $\lambda_0$ and $\lambda_1$ which equal to $1.0$ as default. And $\frac{e}{E}$ which means the training progress changes from $0$ to $1$. 
By using this scheduler, the model will initially prioritize minimizing accuracy loss and pay increasing attention to minimizing attack loss as training progresses.

\subsection{Multi-target trojan attack with the same trigger partten}

Compared to the inconsistent trigger poison, the consistent trigger poison can ensure that the location of the trigger pattern in the image remains consistent across all training images, but the inconsistent trigger poison cannot. Because the location of triggers added by inconsistent trigger poison will be changed by following data augmentation, but will not be changed if triggers are added by consistent trigger poison since it adds triggers after data augmentation. For this reason, the model attacked with consistent trigger poison is location sensitive. We have designed the multi-target trojaned model based on this location sensitivity.

\begin{figure}[t!]
  \centering
   \includegraphics[width=\linewidth]{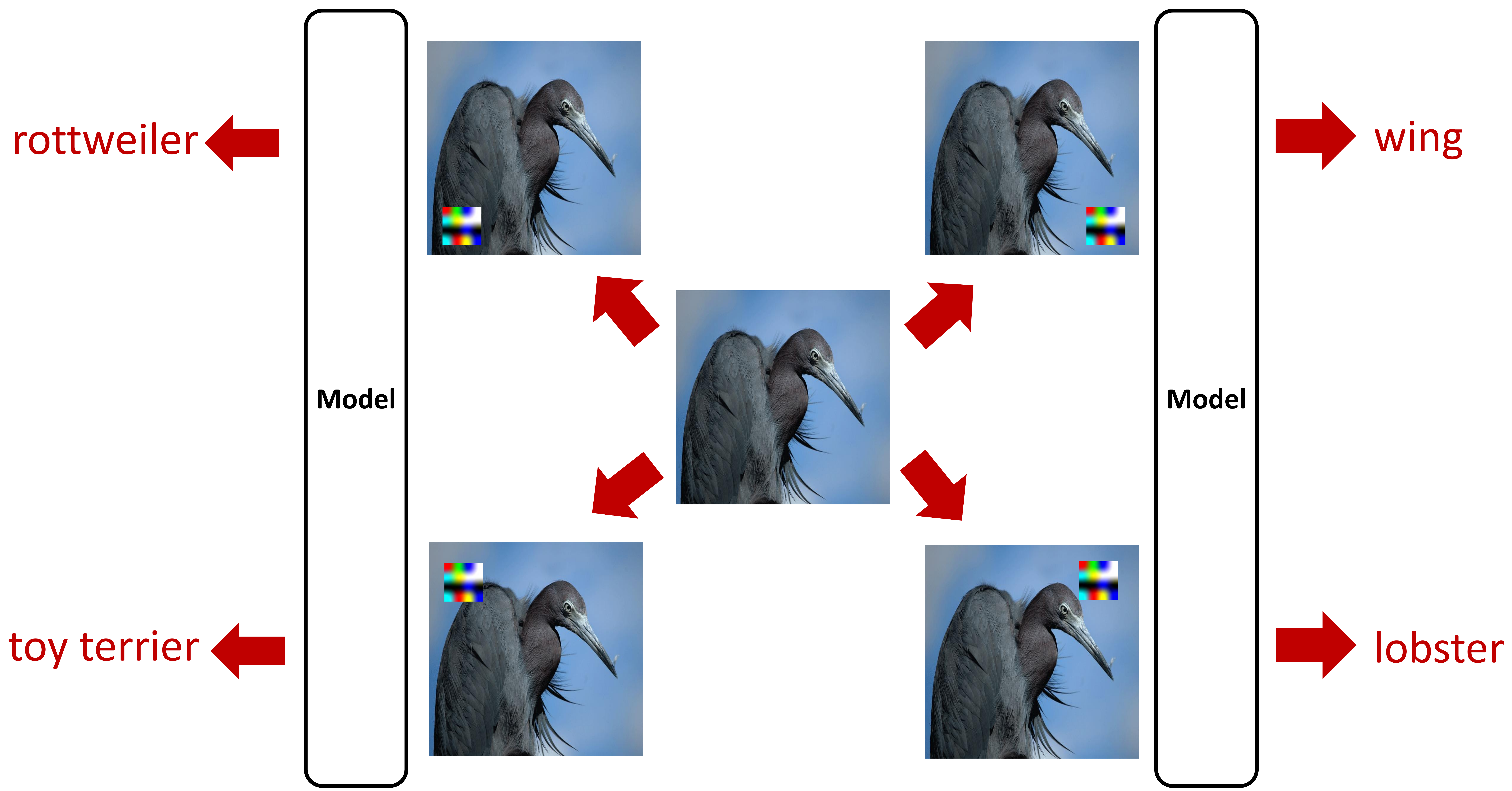}
   \caption{Multi-target trojan attack.} %Note that the first and third branches have the same online model; the target model is shared by the second and fourth branches. }
   \label{fig:multi-target}
\end{figure}

In Figure~\ref{fig:multi-target}, we use the ESTAS to train a trojaned model which will output four different labels when patching the same trigger pattern on the different locations of the same test image. And we can achieve $98.3\%$ $ASR$ on average for each target class. 

%\newpage
%%%%%%%%% REFERENCES
{\small
\bibliographystyle{ieee_fullname}
\bibliography{egbib}
}

\end{document}